\documentclass{svproc}
\usepackage[utf8]{inputenc}
\usepackage{psfrag}
\usepackage[utf8]{inputenc} 
\usepackage{hyperref}       
\usepackage{url}            
\usepackage{amsfonts}       
\usepackage{amsmath} 
\usepackage{amssymb}  
\usepackage{float}
\usepackage{epsfig} 
\usepackage{textcomp}
\usepackage{mathtools}
\usepackage{color}
\usepackage{placeins}
\usepackage{tabularx}
\usepackage{dsfont}
\usepackage{graphicx}
\usepackage{caption}  
\usepackage{float}

\usepackage{algorithm}
\usepackage{listings}
\usepackage{algpseudocode}

\renewcommand{\S}{\mathcal{S}}
\renewcommand{\P}{\mathrm{P}}
\newcommand{\A}{\mathcal{A}}
\newcommand{\R}{\mathcal{R}}

\DeclareMathOperator*{\argmax}{argmax} 

\usepackage{url}

\begin{document}

\mainmatter              
\title{Driving Reinforcement Learning with Models}

\author{Meghana Rathi\inst{1} \and Pietro Ferraro\inst{2}
Giovanni Russo\inst{3}}

\authorrunning{Rathi et al.} 

\tocauthor{Meghana Rathi, Pietro Ferraro and Giovanni Russo}

\institute{School of Electrical \& Electronic Eng. University College Dublin Belfield, Dublin, Ireland \\
\email{meghana.rathi@ucdconnect.ie}
\and
Dyson School of Design Engineering Imperial College London South Kensington, London, UK\\
\email{p.ferraro@imperial.ac.uk} 
\and
Department of Information and Electrical Engineering and Applied Mathematics, Universita' degli Studi di Salerno, Fisciano, Salerno, Italy\\
\email{} WWW home page:
\texttt{http://sites.google.com/view/giovanni-russo/home} }

\maketitle              
\begin{abstract}
In this paper we propose a new approach to complement reinforcement learning (RL) with model-based control (in particular, Model Predictive Control - MPC). We introduce an algorithm, the MPC augmented RL (MPRL) that combines RL and MPC in a novel way so that they can augment each other's strengths. We demonstrate the effectiveness of the MPRL by letting it play against the Atari game Pong. For this task, the results highlight how MPRL is able to outperform both RL and MPC when these are used individually.
\keywords{Model Predictive Control(MPC), Reinforcement Learning(RL)}
\end{abstract}
\section{Introduction}
Model-free reinforcement learning (RL in what follows) has become a popular paradigm to design autonomous agents \cite{8630008,8760436,8772088,8604070}. Its key idea is that of learning a policy for a given task by interacting with the environment via a {\em trial and error} mechanism: essentially, the optimal policy is achieved by exploring the state space and by learning which actions are the best (based on some \emph{reward function}) for a given state.  Unfortunately, two key practical disadvantages of RL are its sample inefficiency and its lack of (e.g. safety) guarantees while learning \cite{4,8458422}.  This paper proposes an approach to drive the learning of RL that stems from the following  observation: in many applications, such as applications requiring physical interactions between the agent and its environment, while a {\em full} model of the environment might not be available, at least {\em parts} of the model, for e.g. a subset of the state space, might be known/identifiable. Motivated by this, we explore how the availability of these {\em partial} models can be leveraged to {\em drive}, and indeed accelerate, the learning phase of RL. 

 \subsection{Contributions}
We propose a novel algorithm that complements the capabilities of RL with those of a well known and established model-based control method, Model Predictive Control (MPC). Our algorithm, the MPC augmented RL (MPRL in what follows), combines RL and MPC in a novel way so that they can augment each other's strengths. The MPRL is inspired by the fact that, from a designer's perspective, complex tasks can be often fulfilled by combining together a set of  {\em functionalities} and, for some of these functionalities, either a mathematical model is known or it might be worth devising it. For example, functionalities for which it is worth to {\em invest} to build a model are these that are critical to satisfy safety requirements: in e.g. an automated driving context, these {\em critical functionalities} are those directly associated to the prevention of crashes and the braking dynamics of the car can be modeled via differential/difference equations. Given these considerations, the MPRL can be described as follows. At each iteration, it checks whether the environment is in a state for which a mathematical model is available (or can be identified). If this is the case, then the action that the agent will apply is computed by leveraging the model and using MPC to optimize a given cost function. If a model is not available, then MPRL makes use of RL (in particular, we will use Q-Learning) to compute a policy from data. The two counterparts (or components) of the MPRL, i.e. MPC and RL, are interlinked and interact within the algorithm. In particular, MPC both drives the state-space exploration and tunes the RL rewards in order to speed the learning process. To the best of our knowledge, this is a new approach to combine MPC and RL, which is complementary to the recent results on learning MPC \cite{8039204,thananjeyan2019safety,doi:10.1146/annurev-control-060117-105215}. A further exploration of the use of feedback control to enhance the performance of data-driven algorithms can also be found in \cite{lellis2019controltutored}, which proposes a different mechanism to complement RL with a feedback control loop. In order to illustrate the effectiveness of MPRL, we let it play the {\em Atari} game {\em Pong} and also control an inverted pendulum. The results highlight the ability of the algorithm to learn the task, outperforming both RL and MPC when these are used individually. The code of our experiments is available at the repository \href{https://github.com/GIOVRUSSO/Control-Group-Code}{https://github.com/GIOVRUSSO/Control-Group-Code}.

\subsection{Related Work}
We now briefly survey some related research threads. 

\noindent{\bf Physics simulation.} The idea of using models and simulation environments to develop intelligent reinforcement learning agents has recently been attracting much research attention, see e.g. \cite{8416737,8648230}. For example, in \cite{6} it is shown how a physical simulator can be embedded in a deep network, enabling agents to both learn parameters of the environment and improve the control performance of the agent. Essentially, this is done by simulating rigid body dynamics via a linear complementarity problem (LCP) technique \cite{7}, \cite{8},  \cite{6}, \cite{12}, \cite{13}. LCP techniques are also used within other simulation environments such as  MuJoCo \cite{9}, Bullet \cite{10}, and DART \cite{11}. Instead, a complementary body of literature investigates the possibility of integrating into networks the mechanisms inspired by the intuitive human ability to understand physics (see e.g. \cite{14}, \cite{15}, \cite{16}, which leverage the ideas of \cite{18}, \cite{19}).

\noindent{\bf Model-based RL.} Although model-free methods, have achieved considerable successes in the recent years, many works suggest that a model-based approach can potentially achieve better performance \cite{4}, \cite{Atkeson97acomparison}, \cite{kurutach2018modelensemble}. The research in model-based RL is a very active area and it is focused on two main settings. The first one makes use of neural networks and a suitable loss function to simulate the dynamics of interest,  whereas another approach makes use of more classical mathematical models  closely resembling system identification \cite{23}. 

\noindent{\bf Safe RL.} The design of safe RL algorithms is a key research topic and different definitions of safety have been proposed in the literature (see e.g. \cite{24}, \cite{25} and references therein). For example, in \cite{DBLP:journals/corr/abs-1109-2147} the authors relate safety to a set of \emph{error states} associated to dangerous/risky situations while in \cite{26} risk adversion is specified in the reward.  A complementary approach is the one of model-based RL where safety is formalized via state space constraints, see e.g. \cite{8651519}. Examples of this approach include \cite{2,4}, where Lyapunov functions are used to show forward invariance of the safety set defined by the constraints. Finally, we note that other approaches include \cite{Garcia:2012:SES:2444851.2444864}, where a priori knowledge of the system is used, in order to craft safe backup policies or \cite{safe} in which authors consider uncertain systems and enforce probabilistic guarantees on their performance.

\section{Background}\label{sec:background}
We now outline the two building blocks composing MPRL.

\noindent{\bf Model Predictive Control.} MPC is a model-based control technique. Essentially, at each time-step, the algorithm computes a control action by solving an optimization problem having as constraint the dynamics of the system being controlled. In addition to the dynamics, other system requirements (e.g. safety or feasibility requirements) can also be formalized as constraints of the optimization problem \cite{GARCIA1989335}. Let $x_k\in\mathbb{R}^n$ be the state variable of the system at time $k$, $u_k\in\mathbb{R}^m$ be its control input and $\eta_k$ be some noise. In this letter we consider discrete-time dynamical systems of the form $x_{k+1} = A_kx_k +D_k+ C_ku_k+\eta_k$ with initial condition $x_{initial}$ and where the time-varying matrices have appropriate dimensions. For this system, formally the MPC algorithm generates the control input $u_k$ by solving the problem
\begin{equation}\label{eqn:MPC}
\begin{split}
&\underset{x_{0:T}\in\mathcal{X},u_{0:T}\in\mathcal{A}}{\text{argmin}}\mathbb{E}\left\{\sum_{t=0}^TJ_t(x_t,u_t)\right\}   \\
& \text{s.t.} \ \ \ \ x_{t+1} = A_tx_t + D_t+ C_tu_t+\eta_t,  \ \ x_0 =x_{initial},
\end{split}
\end{equation}
with $x_{0:T}$ ($u_{0:T}$) denoting the sequence $\{x_0,\ldots,x_T\}$ (resp. $\{u_0,\ldots,u_T\}$) and where: (i) $\mathcal{A}$ and $\mathcal{X}$ are sets modelling the constraints for the valid control actions and states; (ii) $\mathbb{E}\left\{\sum_{t=0}^T J_t(x_t,u_t)\right\}$ is the cost function being optimized, i.e. the expected value of $\sum_{t=0}^T J_t(x_t,u_t)$; (iii) $\eta_t$ is a zero-mean white noise with constant and bounded variance.  See e.g. \cite{Borrelli:2017:PCL:3164811} for more details.

\noindent{\bf Q-Learning and Markov Decision Processes.} Q-Learning (Q-L) is a model free RL algorithm, whose aim is to find an optimal policy with respect to a finite Markov Decision Process (MDP). We adopt the standard formalism for MDPs. A MDP \cite{1} is a discrete stochastic model defined by a tuple $\langle  \S, \A, P, \gamma, \R\rangle$, where: (i) $\S$ is the set of states $s \in \S$; (ii) $\A$ is the set of actions $a \in \A$; (iii) $\P(s'|s,a)$ is the probability of transitioning from state $s$ to state $s'$ under action $a$ ; (iv) $\gamma \in [0,1)$ is the discount factor; (v) $\R(s,a)$ is the reward of choosing the action $a$ in the state $s$. Upon performing an action, the agent receives the reward $\R(s_t,a_t)$. A policy, $\pi$, specifies (for each state) the action that the agent will take and the goal of the agent is that of finding the policy that maximizes the expected discounted total reward. The value $Q^\pi(s,a)$, named Q-function, corresponding to the pair $(s,a)$ represents the estimated expected future reward that can be obtained from $(s,a)$ when using policy $\pi$. The objective of Q-learning is to estimate the Q-function for the optimal policy $\pi^*$, $ Q^{\pi^*}(s,a)$. Define the estimate as $Q(s,a)$. The Q-learning algorithm works then as follows: after setting the initial values for the Q-function,  at each time step, observe current state $s_t$ and select action $a_t$, according to policy $\pi(s_t)$. After receiving the reward $R(s_t,a_t)$ update the corresponding value of the Q-function as follows:
\begin{equation}
\label{eqn:Q_table}
Q(s_t,a_t)\gets (1-\alpha)Q(s_t,a_t) +\alpha\left[ \R(s_t,a_t) + \gamma\max_aQ(s_{t+1},a)\right],
\end{equation}
where $\alpha \in [0,1]$ is called the learning rate. Notice that the Q-learning algorithm does not specify which policy $\pi(\cdot)$ should be considered. In theory, to converge the agent should try every possible action for every possible state many times. For practical reasons, a popular choice for the Q-learning policy is the $\epsilon$-greedy policy, which selects its highest valued (greedy) action, $\pi_\epsilon(s_t) = \argmax_{a_t} Q(s_t,a_t)$, with probability $1-\epsilon(k-1)/k$ and randomly selects among all other $k$ actions with probability $\epsilon/k$ \cite{wunder2010classes}. 
\section{The MPRL Algorithm}
We are now ready to introduce the MPRL algorithm, the key steps of which are summarized as pseudo-code in Algorithm \ref{alg:MPRL}. The main intuition behind this algorithm is that, in many real world systems, tasks can be broken down into a set of functionalities and, for some of these, a mathematical model might be available. Given this set-up, the MPRL aims at combining the strengths of MPC and Q-L. Indeed: (i) within MPRL, MPC can directly control the agent whenever a model is available and, at the same time, it drives the state exploration of Q-L and adjusts its rewards; (ii) on the other hand, Q-L generates actions whenever no mathematical model is available and hence classic model-based control algorithm could not be used.

\noindent The algorithm takes as input the following design parameters: (i) the set of allowed actions, $\A$; (ii) the time horizon and cost function used in (\ref{eqn:MPC}); (iii) the constants $\overline{r}$ and $\underline{r}$, used by MPC to fine tune the reward of Q-L; (iv) an initial matrix $Q(s,a)$. Then, following Algorithm \ref{alg:MPRL}  the following steps are performed:
\begin{description}
\item[{\bf S1:}] at each time-step, MPRL checks whether a model is available. As we will see in Section \ref{sec:application}, for the Pong game, a model can be identified when MPRL is defending against attacks. Instead, in Section \ref{sec:pendulum}, for an inverted pendulum, we use a predefined action around certain operating conditions of the pendulum arm;
\item[{\bf S2a:}] if a model is available, then the action applied by the agent is generated via MPC. Even if the action applied by the agent is given by MPC, the action that would have been obtained via Q-L is also computed. This is done to enable MPC to drive RL. Indeed, if the action from MPC and Q-L are the same, then the $Q(s,a)$ matrix is updated by using the positive reward $\overline{r}$. On the other hand, if the actions from MPC and Q-L differ from one another, then the $Q(s,a)$ matrix is updated with a non-positive reward $\underline{r}$;
\item[{\bf S2b:}] if a model is not available (or cannot be identified), then the agent's action is generated by Q-L;
\item[{\bf S3:}] all relevant quantities are saved within the main algorithm loop.
\end{description}

\begin{algorithm}
\caption{MPRL Algorithm}
\begin{algorithmic}[1]
	\State {\bf Inputs:}
	\State Allowed actions, $\A$
	\State Time horizon, $T$, and cost function $\sum_{t=0}^TJ_t(x_t,u_t)$
	\State Constants $\overline{r}$ and $\underline{r}$
    \State Initial matrix $Q(s,a)$
    \State{\bf Main loop:}
	\For{$k=0,\ldots$}
	\State Check if the model in (\ref{eqn:MPC}) is known or can be identified
	\If{Model is available}
		\State Get $s_k$ and $x_k$
		\If{$k\ge1$}
		\If{$a_{k-1}= u_{k-1}$}
	\State $Q(s_{k-1},u_{k-1})\gets (1-\alpha)Q(s_{k-1},u_{k-1})+\alpha\left[\overline{r}+\gamma\max_a\{Q(s_k,a)\}\right]$
	\Else
		\State $Q(s_{k-1},u_{k-1})\gets (1-\alpha)Q(s_{k-1},u_{k-1})+\alpha\left[\underline{r}+\gamma\max_a\{Q(s_k,a)\}\right]$
	\EndIf
	\EndIf
	\State $x_{init} \gets x_k$ in (\ref{eqn:MPC})
	\State Compute $u_k$ via MPC  and $a_k$ using Q-L 
	\State Apply $u_k$
	\State Save $x_k$, $s_k$, $a_k$, $u_k$
	\Else
	\State Apply $a_k$ computed via Q-L
	\State Save $s_k$, $a_k$
	\EndIf
	\State $k\gets k+1$
	\EndFor

\end{algorithmic}\label{alg:MPRL}
\end{algorithm}

\section{Using MPRL to learn Pong}\label{sec:application}
We now illustrate the effectiveness of MPRL by letting it play against the Atari game {\em Pong}. Pong is a 2 player game where each player moves a paddle in order to bounce a ball to the opponent. A player scores a point when the opponent fails to bounce the ball back and the game ends when a player scores $21$ points. In what follows, we give a thorough description of how Algorithm \ref{alg:MPRL} has been implemented in order to allow MPRL to play against Pong.

\subsection{The environment and data gathering}
\begin{figure}[tb]
\centering
\begin{tabular}{cc} 
{\includegraphics[width=0.5\textwidth]{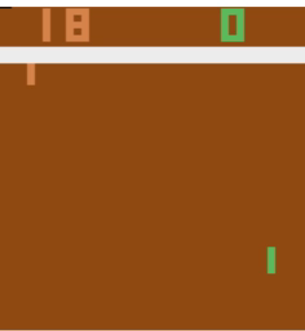}}&
{\includegraphics[width=0.5\textwidth]{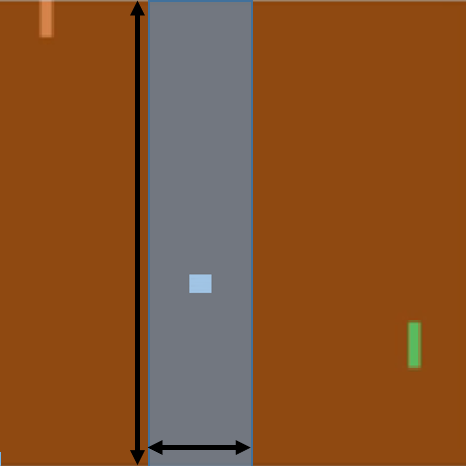}}  
\end{tabular}
\caption{Left panel: a typical frame from Pong. The paddle moved by MPRL is the green one. Right panel: a zoom illustrating the $80\times12$ pixels window used to extract the new position of the ball, given its previous position.}
\label{fig:environment}
\end{figure}

The environment of the game was set-up using the OpenAI gym library in Python. In particular, we used the {\em PongDeterministic-v4} (with $4$ frame skips) configuration of the environment, which is the one used to assess Deep Q-Networks, see e.g. \cite{mnih-atari-2013}. The configuration used has, as observation space, $\text{Box}(210,160,3)$ (see Figure \ref{fig:environment}, left panel)\footnote{See http://gym.openai.com/docs/ for documentation on the environment observation space}. Within our experiments, we first removed the part of the images that contained the game score and this yielded an observation space of $\text{Box}(160,160,3)$ and then we down-sampled the resulting image to get a {\em reduced} observation space of $\text{Box}(80,80,3)$ so that each frame consists of a matrix of $80\times80$ pixels. Within the image, a coordinate system is defined within the environment, with the origin of the $x$ and $y$ axes being in the bottom-right corner.

\noindent Given the above observation space, both the position of the ball and the vertical position of the paddle moved by MPRL were extracted from each frame. In particular:
\begin{itemize}
\item At the beginning of the game, the centroid of the ball is found by iterating through the frame to find the location of all pixels with a value of $236$ (this corresponds to the white color, i.e. the color of the ball in Pong). Then, once the ball is found the first time, the frame is only scanned in a window around the position of the ball previously found (namely, we used a window of $80\times12$ pixels, see Figure \ref{fig:environment}, right panel);
\item Similarly the paddle's centroid position is found by scanning the frame for pixels having value $92$ (this corresponds to the green color, i.e. the color of the MPRL paddle).
\end{itemize}

\subsection{Definition of the task and its functionalities}
The agent's task is that of winning the game, which essentially consists of two phases: (i) {\em defense} phase, where the agent needs to move the paddle to bounce the ball in order to avoid that the opponent makes a point; (ii) an {\em attack} phase, where the agent needs instead to properly bounce the ball in order to make the point. During the defense phase, MPRL used its MPC component (implemented as described in Section \ref{sec:MPC_impl}) to move the paddle. Indeed, this phase is completely governed by the {\em physics} of the game and by the moves of our agent. This, in turn, makes it possible to identify, from pixels, both the ball and the paddle dynamics (Section \ref{sec:MPC_impl}). Instead, we used the Q-L component of MPRL during the attack phase. Indeed, even if a mathematical model describing the evolution of the position of the ball could be devised, there is no difference equation that could predict what our opponent would do in response (as we have no control over it). Therefore, in the attack phase, we let the Q-L counterpart of our algorithm learn how to score a point.
\subsection{Implementing MPC}\label{sec:MPC_impl}
We describe the MPC implementation used within MPRL to play Pong by first introducing the mathematical model serving as the constraint in (\ref{eqn:MPC}). This model describes both the dynamics of the ball and of the paddle moved by MPRL.

\noindent{\bf Ball dynamics.} We denote by $x^{(b)}_t$ and $y^{(b)}_t$ the $x$ and $y$ coordinates of the centroid of the ball at time $t$. The mathematical model describing the dynamics of the ball is then:
\begin{equation}\label{eqn:MPC_model_1}
\begin{split}
x^{(b)}_{t+1} = x^{(b)}_t + v_{t,x}, \hspace{5mm} y^{(b)}_{t+1} = y^{(b)}_t + v_{t,y},
\end{split}
\end{equation}
where $x^{(b)}_{t+1}$ and $y^{(b)}_{t+1}$ are the predicted next coordinates at time $t+1$ and where the speeds at time $t$, i.e. $v_{t,x}$ and $v_{t,y}$ are computed from the positions extracted from the current and the two previous frames, i.e. $x^{(b)}_{t-2},x^{(b)}_{t-1},x^{(b)}_t$ and $y^{(b)}_{t-2},y^{(b)}_{t-1},y^{(b)}_t$. In particular, this is done by first computing the quantities $v_{x1},v_{x2}$ and $v_{y1},v_{y2}$ as follows:
\begin{equation}\label{eqn:MPC_int}
\begin{bmatrix}
    v_{x_{1}}\\
    v_{y_{1}}
\end{bmatrix}= \begin{bmatrix}
    x^{(b)}_{t}\\
    y^{(b)}_{t}
\end{bmatrix}
-
\begin{bmatrix}
    x^{(b)}_{t-1}\\
    y^{(b)}_{t-1}
\end{bmatrix},
\ \ \ 
\begin{bmatrix}
    v_{x_{2}}\\
    v_{y_{2}}
\end{bmatrix}= \begin{bmatrix}
    x^{(b)}_{t-1}\\
    y^{(b)}_{t-1}
\end{bmatrix}
-
\begin{bmatrix}
    x^{(b)}_{t-2}\\
    y^{(b)}_{t-2}
\end{bmatrix}.
\end{equation}
Consider now the speed along the $x$ axis. If there is no impact between the ball and the paddle, we set $v_{t,x}= 0.5(v_{x1}+v_{x2}) +var\left([v_{x1},v_{x2}]\right)=\bar v_{t,x}+\eta_{t,x}$ (where $var(a)$ denotes the variance of the generic vector $a$ and $\eta_{t,x}$ is a white noise with zero mean and variance $var\left([v_{x1},v_{x2}]\right)$). Instead, along the $y$ axis, we have $v_{t,y}= 0.5(v_{y1}+v_{y2}) +var\left([v_{y1},v_{y2}]\right)=\bar v_{t,x}+\eta_{t,x}$ (with $\eta_{t,y}$ being a white noise with zero mean and variance $var\left([v_{y1},v_{y2}]\right)$) if there has been no impact and $v_{t,x}=v_{x1}$ if there has been an impact of the ball with one of the walls. 

\noindent{\bf Paddle dynamics.} In the gym environment used for the experiments, the only control action that can be applied by MPRL at time $t$, i.e. $u_t$ is that of moving its paddle. In particular, the agent can either move the paddle up ($u_t=1$) or down ($u_t=-1$) or simply not moving the paddle ($u_t=0$). It follows that, given the vertical position of the centroid of the paddle at time $t$, say $y^{(p)}_t$, its dynamics can be modeled by
\begin{equation}\label{eqn:MPC_model_2}
\begin{split}
y^{(p)}_{t+1} = y^{(p)}_t + u_t.
\end{split}
\end{equation}

\noindent{\bf The MPC model and the cost function.} Combining the models in (\ref{eqn:MPC_model_1}) - (\ref{eqn:MPC_model_2}) yields the dynamical system serving as constraint in (\ref{eqn:MPC}). Note that the resulting model can be formally written as the system in (1) once the state $x_t$ is defined as $x_t =[x_t^{(b)}, y_t^{(b)},y_t^{(p)}]^T$. Finally, in the implementation of the MPC algorithm, we used as cost function $\sum_{t=0}^TJ_t = \|y^{(b)}_{t+T}-y^{(p)}_{t+T}\|^2$. That is, with this choice of cost function the algorithm seeks to regulate the paddle's position so that the distance between the position of the ball at time $t$ and the position of the MPRL paddle at time $t+T$ is minimised.  Note that the time horizon, $T$, used in the above cost function is obtained by propagating the ball model (\ref{eqn:MPC_model_1}) in order to estimate after how many iterates the ball will hit the border protected by the MPRL paddle.

\subsection{Implementing Q-L}
We implemented the Q-L algorithm outlined in Section \ref{sec:application}. The set of actions available to the agent were $a_t\in\{-1,0,+1\}$, while the state at time $s_t$ was defined as the $5$-dimensional vector containing: (i) the coordinates of the position at time time $t$ of the ball (in pixels); (ii) the velocity of the ball (rounded to the closest integer) across the $x$ and $y$ axes; (iii) the position of the paddle moved by MPRL. The reward was obtained from the game environment: our agent was given a reward of $+1$, each time the opponent missed to hit the ball, and $-1$ each time our agent missed to hit the ball. Finally, the values of the Q-table were initialized to $0$. In the experiments, the state-action pair was updated whenever a point was scored and a greedy policy was used to select the action. Also, in the experiments we set both $\alpha$ and $\gamma$ in (\ref{eqn:Q_table}) to $0.7$.  Moreover, following Algorithm \ref{alg:MPRL}, the Q-function was also updated when our agent was defending (i.e. when MPRL was using MPC to move the paddle). In particular, within the experiments we assigned: (i) a positive reward, $\overline{r}$, whenever the action from Q-L and MPC were the same; (ii) a non-positive reward, $\underline{r}$, whenever the actions were not the same. In this way, within MPRL, the Q-L component is driven to learn defence tactics too.

\subsection{Handover between MPRL components}
\noindent Finally, we now describe how the handover between the MPC and RL components of MPRL was implemented in the experiments. Intuitively, the paddle was moved by the MPC component when the ball was coming towards the MPRL paddle and, at the same time, the future vertical position of the ball (predicted via the model) was far from the actual position of the agent's paddle. That is, MPC was used whenever the following conditions were simultaneously satisfied: $v^{(b)}_{t,x} <0$, $\| y^{(b)}_{t+T} - y^{(p)}_t\| > H_y$. In all the other situations the paddle was moved by actions generated by the Q-L component. Note that: (i) $v^{(b)}_t$ is estimated from the game frames as described above (in the environment negative velocities along the $x$ axis mean that the ball is coming towards the green paddle); (ii) the computation of $y^{(b)}_{t+T}$ relies on simulating the model describing the ball's dynamics (\ref{eqn:MPC_model_1}); (iii) $H_y$ is a threshold and this is a design parameter. 

\subsection{Results}
We are now ready to present the results obtained by letting MPRL play Pong. The results are quantified by plotting the {\em game reward} as a function of the number of {\em episodes} played by MPRL. An episode consists of as many rounds of pong it takes for one of the players to reach $21$ points, while the game reward is defined to be the difference between the points scored by MPRL within the episode  and the points scored by the opponent within the episode. Essentially, a negative game reward means that MPRL lost that episode; the lowest possible value that can be attained is $-21$ and this happens when MPRL is not able to score any point. Viceversa, a positive game reward means that the agent was able to beat the opponent; the maximum value that can be obtained is $+21$, when the opponent did not score any point.

\begin{figure}[htb]
\begin{center}
\centering
\begin{tabular}{cc} 
\includegraphics[trim={0 0 0 60},clip,width=\textwidth/2]{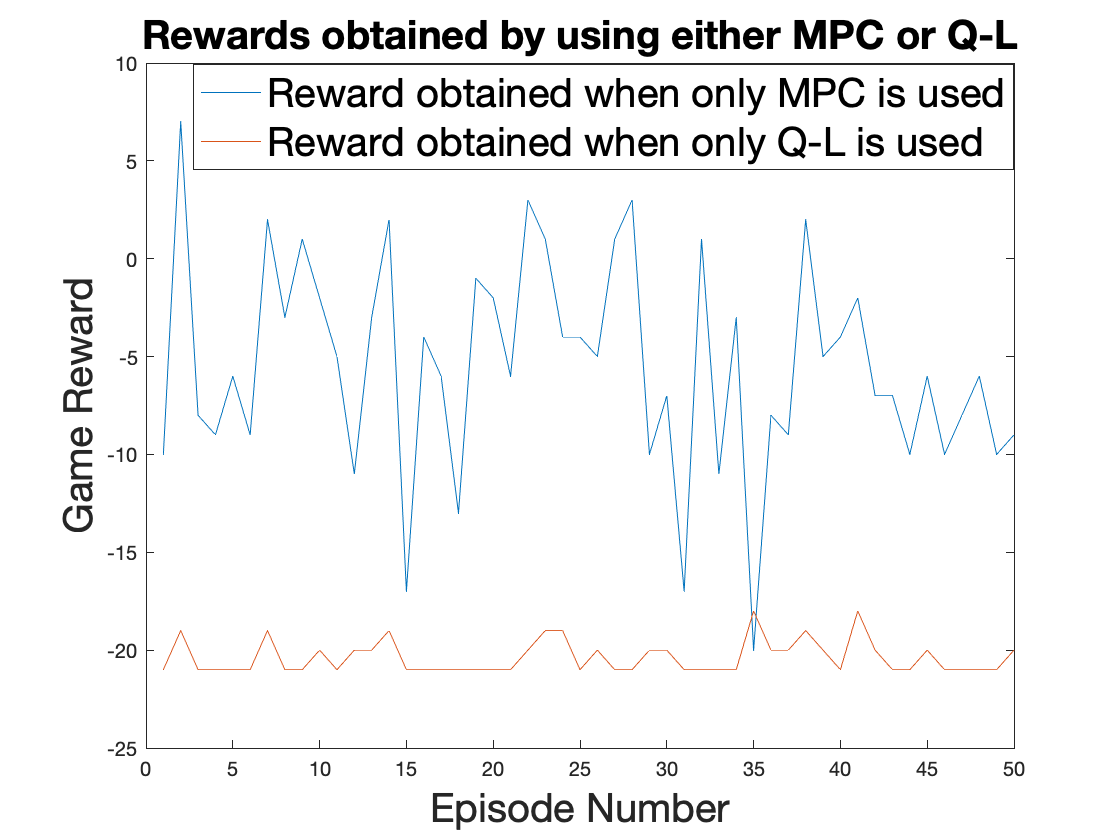}&
\includegraphics[trim={0 0 0 60},clip,width=\textwidth/2]{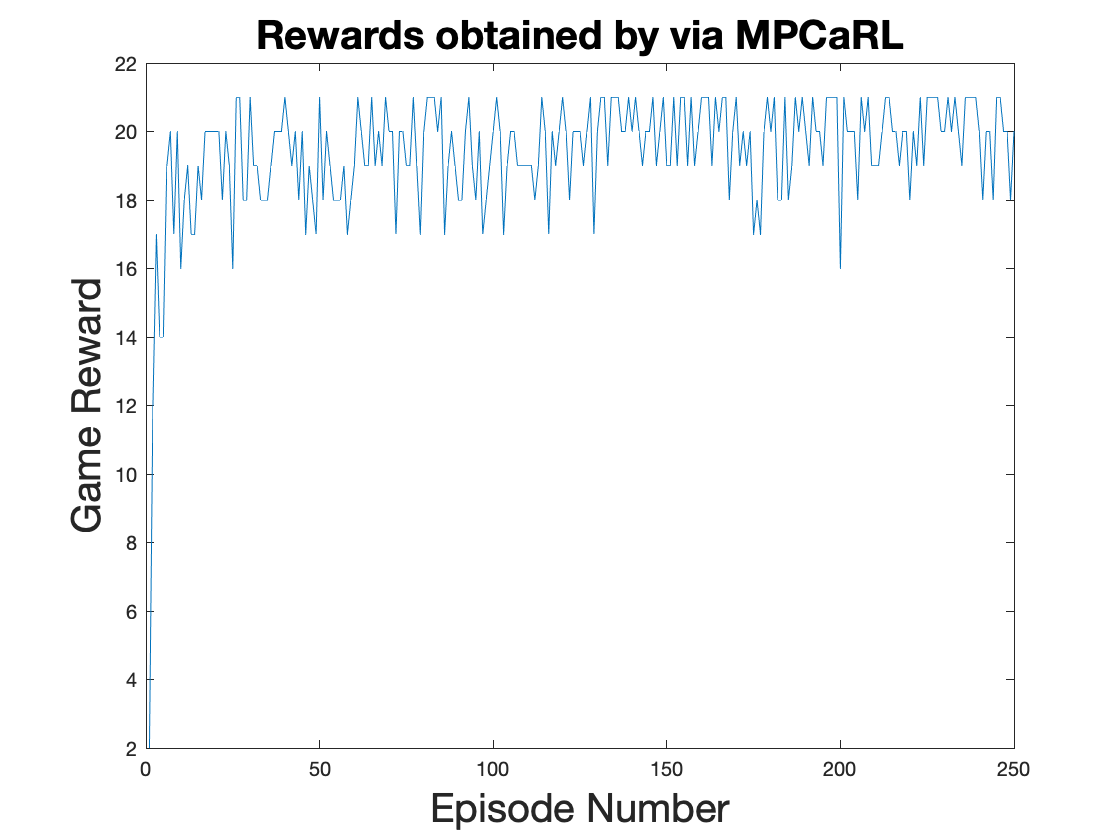}
\end{tabular}
  \caption{Left panel: episode rewards as a function of the number of episodes played when either MPC or Q-L are used. Right panel: episode rewards as a function of the number of episodes played by MPRL. Parameters of MPRL were set as follows: $\overline{r}=0.1$, $\underline{r} =0$, $H_y =5$. }
  \label{fig:experiments_1}
  \end{center}
\end{figure}

\noindent As a first experiment, we implemented an agent that would only use MPC or the Q-L algorithm (without prior training). We let this agent play Pong for $50$ episodes and, as the left panel in Figure \ref{fig:experiments_1} shows, as expected, the Q-L agent did not obtain good rewards in the first $50$ episodes, consistently loosing games with a difference in the scores of about $20$ points. Instead, when the agent used the MPC described in Section \ref{sec:application} better performance were obtained. These performance, however, were not comparable with those obtained via a trained Q-L agent and the reason for this is that, while MPC allows the agent to defend, it does not allow for the learning of an attack strategy to consistently obtain points. Using MPRL allowed to overcome the shortcomings of the MPC and Q-L agents. In particular, as shown in the right panel of Figure \ref{fig:experiments_1}, when the MPRL agent played against Pong, it was able to consistently beat the game (note the agent never lost a game) while quickly learning an attack strategy to obtain high game rewards. Indeed, note how the agent is able to consistently obtain rewards of about $20$ within $50$ episodes.

\begin{figure}[htb]
\begin{center}
\centering
\begin{tabular}{ccc} 
\includegraphics[trim={0 0 0 70},clip,width=\textwidth/3]{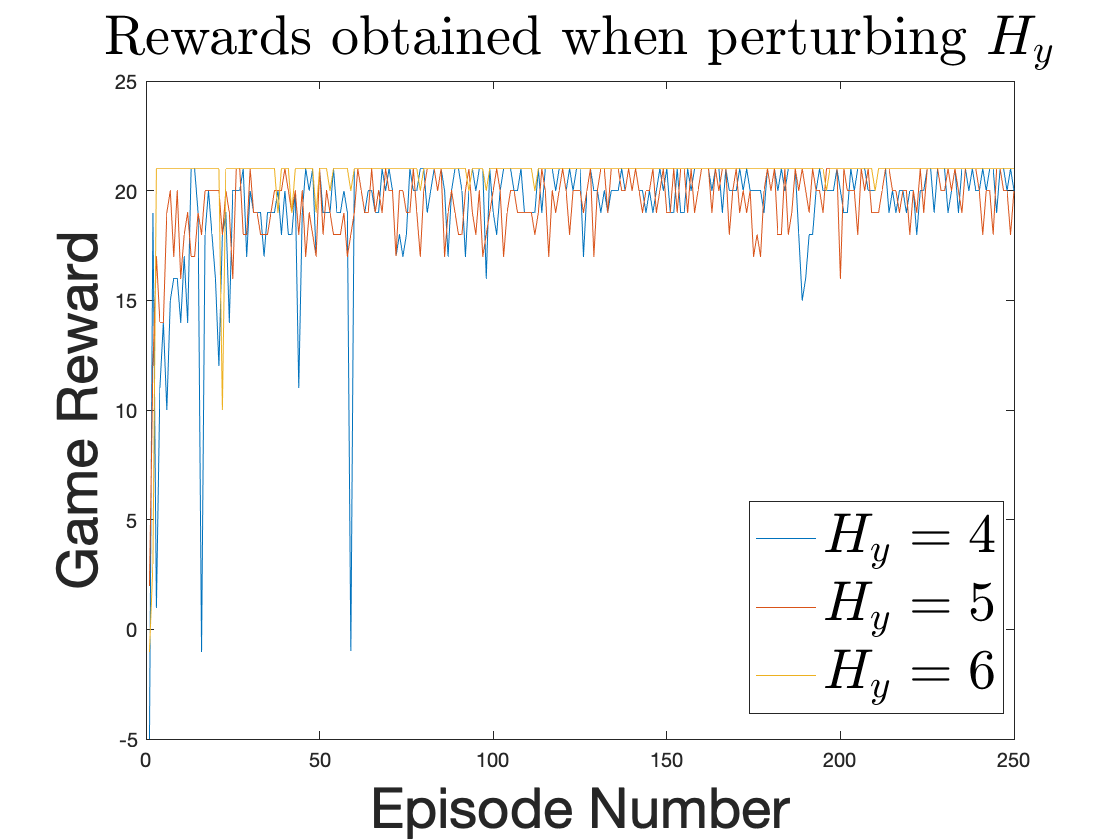}&
\includegraphics[trim={0 0 0 70},clip,width=\textwidth/3]{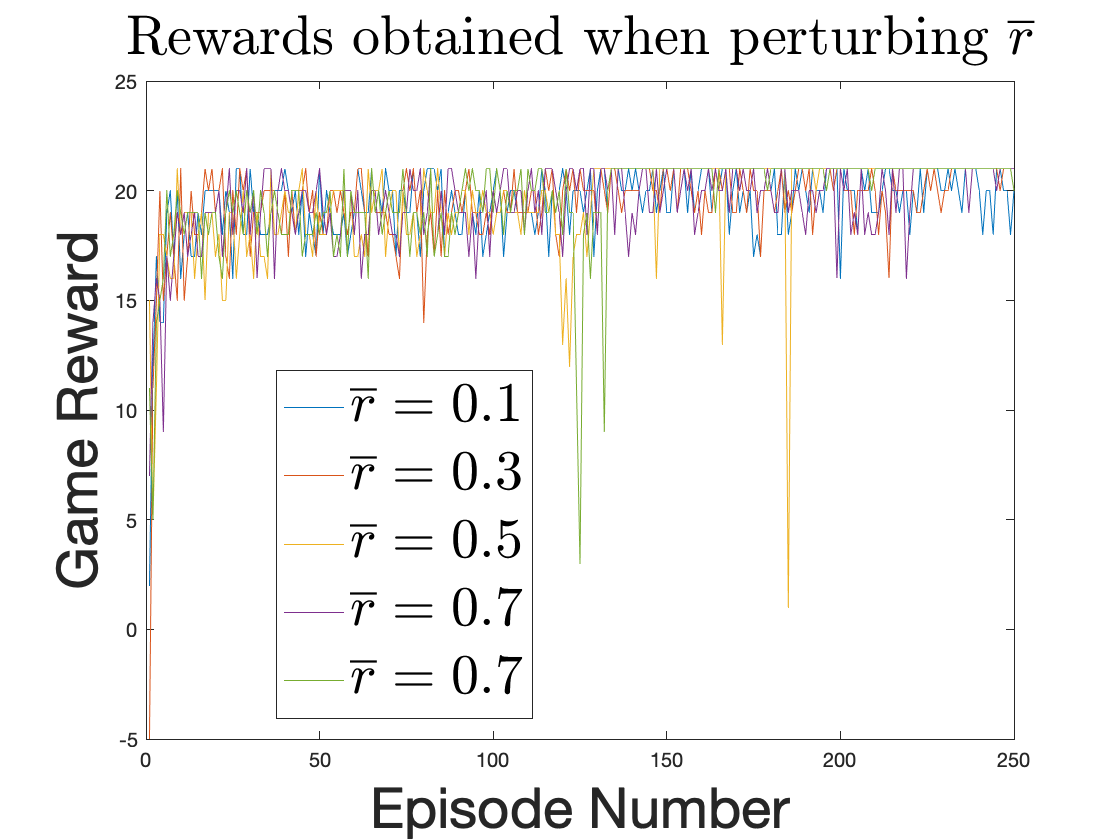}&
\includegraphics[trim={0 0 0 70},clip,width=\textwidth/3]{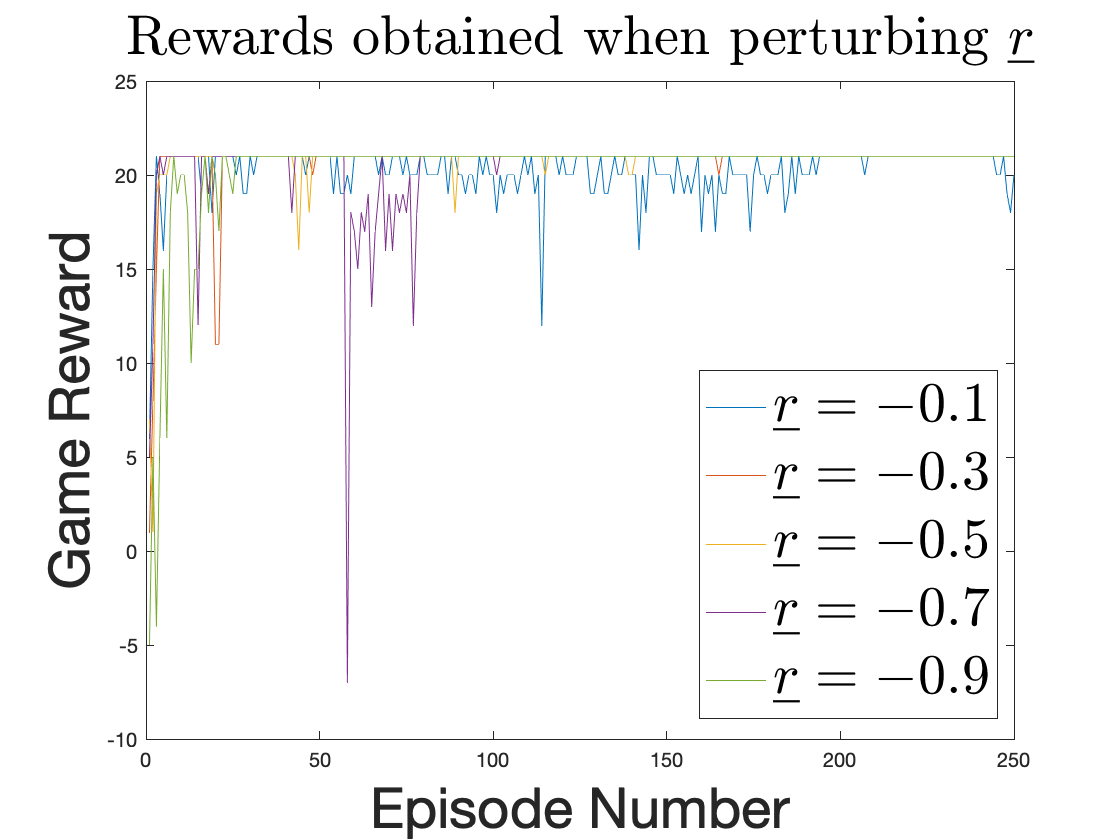}
\end{tabular}
  \caption{Left panel: rewards obtained by the MPRL when $H_y$ is perturbed. All the other parameters were kept unchanged (i.e. $\overline{r}=0.1$, $\underline{r} =0$). Middle panel: rewards obtained by MPRL when $\overline{r}$ is perturbed. In this experiment, $H_y = 5$ and $\underline{r}=0$. Right panel: rewards obtained by MPRL when $\underline{r}$ is perturbed. In the experiment, $H_y =5$ and $\overline{r}=0.1$.}
  \label{fig:variations}
  \end{center}
\end{figure}

\noindent In order to further investigate the performance of MPRL, we also evaluate its performance when the parameters  $\overline{r}$, $\underline{r}$ and $H_y$ are changed. We first take $H_y$ in consideration; $H_y$ determines when the handover between MPC and RL takes place. As Figure \ref{fig:variations} (left panel) illustrates, MPRL is still able to consistently obtain high rewards when $H_y$ is perturbed, i.e. $H_y \in\{4, 5, 6\}$. Notice that when $H_y = 4$, while MPRL is still able to beat the game, it also experiences some drops in the rewards. This phenomenon, which will be further investigated in future studies, might be due to the fact that restricting the space of movements of the Q-L component of the algorithm also restricts its learning capabilities. Instead, when the manoeuvre space of the Q-L is bigger ($H_y = 6$), the algorithm, due to the increased flexibility in the moves, is able to learn better moves faster. As a further experiment, we fixed $H_y = 5$ and perturbed the algorithm parameter $\overline{r}$ so that $\overline{r}\in\{0.1,0.3,0.5,0.7,0.9\}$. The results of this experiment are shown in the middle panel of Figure \ref{fig:variations}. It can be noted that smaller values of $\overline{r}$ (i.e. $\overline{r}= 0.1$ and $\overline{r} = 0.3$) lead to a more consistent performance as compared to the higher values of $\overline{r}$ which lead to negative spikes in the performance. This behaviour might be  due to the fact that higher values of $\overline{r}$ essentially imply that MPRL trusts more the MPC actions than Q-L actions, hence penalizing the ability of MPRL to quickly learn the attack strategy. Intuitively, simulations show that, while the MPC component is important for enhancing the agent's defense, {\em too much influence} of this component on the Q-L can reduce the attack performance of the agent (this, in turn, is essential in order to score higher points). Consistently, the same behaviour can be observed when $\underline{r}$ is perturbed ($H_y = 5$, $\overline{r}=0.1$ and $\underline{r} \in \{-0.1,-0.3,-0.5,-0.7,-0.9 \}$), as shown in Figure \ref{fig:variations} (right panel), where it can be observed that the most negative values $\underline{r}$ lead to dips in performance. 

\section{MPRL to control an inverted pendulum}\label{sec:pendulum}
\begin{figure}[htb]
\begin{center}
\centering
\includegraphics[width=0.6\textwidth]{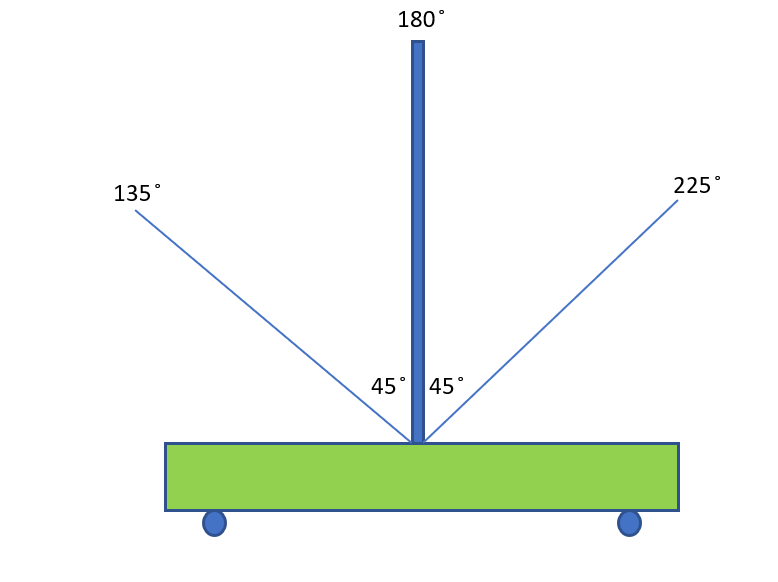}
  \caption{Inverted pendulum in its balanced state within the safety limits at 135 and 225 degrees.}
  \label{fig:balancedpendulum}
  \end{center}
\end{figure}

An additional experiment was carried out  to test the capability of the MPRL algorithm in a different domain. This experiment involved balancing an inverted pendulum in a virtual environment using MPRL and comparing its results against using Q-L only. Specifically, the goal of the MPRL was to  move the cart by controlling its speed so that the pendulum would stay in its upright position at 180 degrees (see Figure \ref{fig:balancedpendulum}).

\begin{figure}[htb]
\begin{center}
\centering
\includegraphics[width=1\textwidth]{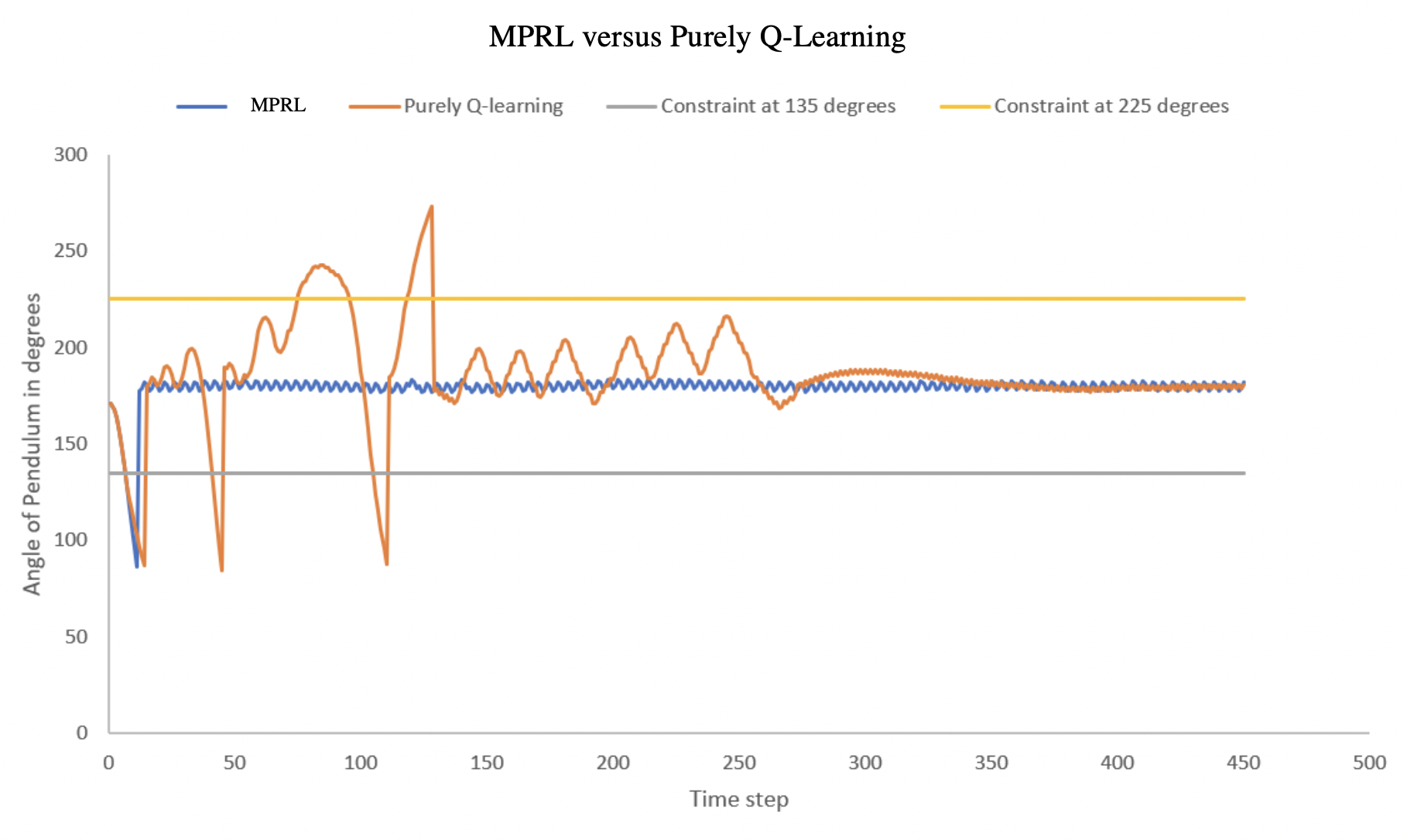}
  \caption{Angle of the inverted pendulum against time when MPRL is used and when Q-learning is used. The upright position is 180 degrees and the constraints at 135 and 225 degrees are the angles past which the pendulum is highly likely to fall over.}
  \label{fig:pendulum}
  \end{center}
\end{figure}

In our experiments, the reward was specified to be +1 if the pendulum was within 175 and 185 degrees for each time step and -1 if the pendulum went outside this range. Moreover, in our experiments we considered the  region between 135 and 225 degrees as a \emph{safe} region and we wanted the MPRL to keep the arm of the pendulum within this region, see Figure \ref{fig:pendulum}. In order to handle this {\em safety} requirement, in our MPRL we used an MPC-like algorithm whenever the pendulum arm was outside the region 135 - 225 degrees. In particular: (i) if the arm position was at 135 degrees (or smaller), then the cart speed was adjusted by MPRL to move the cart to the left; (ii) if the arm position was at 225 degrees or higher, then the cart was moved by MPRL to the right. By doing so, the numerical results showed that {\em MPC component} of MPRL was able to bring the arm within the safety region (i.e. between 135 and 225 degrees). Inside this region, the Q-L component of MPRL was active.

Again, as shown in Figure \ref{fig:pendulum}, the results confirm the capability of the MPRL to quickly learn how to balance the pendulum. Moreover, when compared to the results obtained via Q-Learning, the MPRL dramatically reduces the violation of the {\em safety} constraints (i.e. the number of times when the pendulum is outside the safe region) and learns faster. This again shows that MPC, and more generally models, can augment the performance of purely data-driven techniques.

\section{Conclusions and Future Work}\label{Sec: Conclusions}
We investigated the possibility of combining Q-L and model-based control so that they can augment each other's capabilities. In doing so, we introduced a novel algorithm, the MPRL that: (i) leverages MPC when a mathematical model is available and Q-L otherwise; (ii) uses MPC to both drive the state-space exploration of Q-L and to fin tune its rewards. We first illustrated the effectiveness of our algorithm by letting it play against Pong and by analysing its performance when the algorithm parameters are perturbed. Interestingly, the experiments highlight how the algorithm is able to outperform both Q-L and MPC when these are used individually. Moreover, we also tested MPRL in a different domain and showed how it quickly learns to balance an inverted pendulum, while reducing the number of violations of safety constraints. Our future work will include the implementation of the MPRL in different settings, the study of its convergence properties and the design of optimal handover strategies.
%
%
\bibliographystyle{IEEEtran}
\bibliography{russobib.bib}

\begin{thebibliography}{10}
\providecommand{\url}[1]{#1}
\csname url@samestyle\endcsname
\providecommand{\newblock}{\relax}
\providecommand{\bibinfo}[2]{#2}
\providecommand{\BIBentrySTDinterwordspacing}{\spaceskip=0pt\relax}
\providecommand{\BIBentryALTinterwordstretchfactor}{4}
\providecommand{\BIBentryALTinterwordspacing}{\spaceskip=\fontdimen2\font plus
\BIBentryALTinterwordstretchfactor\fontdimen3\font minus
  \fontdimen4\font\relax}
\providecommand{\BIBforeignlanguage}[2]{{%
\expandafter\ifx\csname l@#1\endcsname\relax
\typeout{** WARNING: IEEEtran.bst: No hyphenation pattern has been}%
\typeout{** loaded for the language `#1'. Using the pattern for}%
\typeout{** the default language instead.}%
\else
\language=\csname l@#1\endcsname
\fi
#2}}
\providecommand{\BIBdecl}{\relax}
\BIBdecl

\bibitem{8630008}
M.~{Breyer}, F.~{Furrer}, T.~{Novkovic}, R.~{Siegwart}, and J.~{Nieto},
  ``Comparing task simplifications to learn closed-loop object picking using
  deep reinforcement learning,'' \emph{IEEE Robotics and Automation Letters},
  vol.~4, no.~2, pp. 1549--1556, 2019.

\bibitem{8760436}
S.~{Hoppe}, Z.~{Lou}, D.~{Hennes}, and M.~{Toussaint}, ``Planning approximate
  exploration trajectories for model-free reinforcement learning in
  contact-rich manipulation,'' \emph{IEEE Robotics and Automation Letters},
  vol.~4, no.~4, pp. 4042--4047, 2019.

\bibitem{8772088}
B.~{Liu}, L.~{Wang}, and M.~{Liu}, ``Lifelong federated reinforcement learning:
  A learning architecture for navigation in cloud robotic systems,'' \emph{IEEE
  Robotics and Automation Letters}, vol.~4, no.~4, pp. 4555--4562, 2019.

\bibitem{8604070}
X.~{Tan}, C.~{Chng}, Y.~{Su}, K.~{Lim}, and C.~{Chui}, ``Robot-assisted
  training in laparoscopy using deep reinforcement learning,'' \emph{IEEE
  Robotics and Automation Letters}, vol.~4, no.~2, pp. 485--492, 2019.

\bibitem{4}
F.~Berkenkamp, M.~Turchetta, A.~Schoellig, and A.~Krause, ``Safe model-based
  reinforcement learning with stability guarantees,'' \emph{Advances in Neural
  Information Processing Systems 30}, pp. 908--918, 2017.

\bibitem{8458422}
M.~{Pfeiffer}, S.~{Shukla}, M.~{Turchetta}, C.~{Cadena}, A.~{Krause},
  R.~{Siegwart}, and J.~{Nieto}, ``Reinforced imitation: Sample efficient deep
  reinforcement learning for mapless navigation by leveraging prior
  demonstrations,'' \emph{IEEE Robotics and Automation Letters}, vol.~3, no.~4,
  pp. 4423--4430, Oct 2018.

\bibitem{8039204}
U.~{Rosolia} and F.~{Borrelli}, ``Learning model predictive control for
  iterative tasks. a data-driven control framework,'' \emph{IEEE Transactions
  on Automatic Control}, vol.~63, no.~7, pp. 1883--1896, July 2018.

\bibitem{thananjeyan2019safety}
B.~Thananjeyan, A.~Balakrishna, U.~Rosolia, F.~Li, R.~McAllister, J.~E.
  Gonzalez, S.~Levine, F.~Borrelli, and K.~Goldberg, ``Safety augmented value
  estimation from demonstrations (saved): Safe deep model-based rl for sparse
  cost robotic tasks,'' 2019.

\bibitem{doi:10.1146/annurev-control-060117-105215}
U.~Rosolia, X.~Zhang, and F.~Borrelli, ``Data-driven predictive control for
  autonomous systems,'' \emph{Annual Review of Control, Robotics, and
  Autonomous Systems}, vol.~1, no.~1, pp. 259--286, 2018.

\bibitem{lellis2019controltutored}
F.~D. Lellis, F.~Auletta, G.~Russo, and M.~di~Bernardo, ``Control-tutored
  reinforcement learning: an application to the herding problem,'' 2019.

\bibitem{8416737}
M.~{Pecka}, K.~{Zimmermann}, M.~{Petrlík}, and T.~{Svoboda}, ``Data-driven
  policy transfer with imprecise perception simulation,'' \emph{IEEE Robotics
  and Automation Letters}, vol.~3, no.~4, pp. 3916--3921, 2018.

\bibitem{8648230}
M.~{Hazara} and V.~{Kyrki}, ``Transferring generalizable motor primitives from
  simulation to real world,'' \emph{IEEE Robotics and Automation Letters},
  vol.~4, no.~2, pp. 2172--2179, 2019.

\bibitem{6}
F.~de~Avila Belbute-Peres, K.~Smith, K.~Allen, J.~Tenenbaum, and J.~Z. Kolter,
  ``End-to-end differentiable physics for learning and control,''
  \emph{Advances in Neural Information Processing Systems 31}, pp. 7178--7189,
  2018.

\bibitem{7}
R.~W. Cottle, ``Linear complementarity problem,'' in \emph{Encyclopedia of
  Optimization}, C.~A. Floudas and P.~M. Pardalos, Eds.\hskip 1em plus 0.5em
  minus 0.4em\relax Boston, MA: Springer US, 2009, pp. 1873--1878.

\bibitem{8}
M.~B. Cline, ``Rigid body simulation with contact and constraints,'' Ph.D.
  dissertation, 2002.

\bibitem{12}
J.~Degrave, M.~Hermans, J.~Dambre, and F.~wyffels, ``A differentiable physics
  engine for deep learning in robotics,'' \emph{Frontiers in Neurorobotics},
  vol.~13, p.~6, 2019.

\bibitem{13}
A.~Lerer, S.~Gross, and R.~Fergus, ``Learning physical intuition of block
  towers by example,'' in \emph{33rd International Conference on Machine
  Learning}, vol.~48, 2016, pp. 430--438.

\bibitem{9}
E.~{Todorov}, T.~{Erez}, and Y.~{Tassa}, ``{M}u{J}o{C}o: a physics engine for
  model-based control,'' in \emph{2012 IEEE/RSJ International Conference on
  Intelligent Robots and Systems}, Oct 2012, pp. 5026--5033.

\bibitem{10}
J.~Lee, M.~X. Grey, S.~Ha, T.~Kunz, S.~Jain, Y.~Ye, S.~S. Srinivasa,
  M.~Stilman, and C.~K. Liu, ``{DART}: Dynamic animation and robotics
  toolkit,'' \emph{The Journal of Open Source Software}, vol.~3, no.~22, p.
  500, Feb 2018.

\bibitem{11}
M.~Hermans, B.~Schrauwen, P.~Bienstman, and J.~Dambre, ``Automated design of
  complex dynamic systems,'' \emph{PLOS ONE}, vol.~9, no.~1, pp. 1--11, 01
  2014.

\bibitem{14}
M.~B. Chang, T.~Ullman, A.~Torralba, and J.~B. Tenenbaum, ``A compositional
  object-based approach to learning physical dynamics,'' in \emph{5th
  International Conference on Learning Representations}, 2017.

\bibitem{15}
P.~Battaglia, R.~Pascanu, M.~Lai, D.~J. Rezende, and K.~Kavukcuoglu,
  ``Interaction networks for learning about objects, relations and physics,''
  in \emph{Proceedings of the 30th International Conference on Neural
  Information Processing Systems}, 2016, pp. 4509--4517.

\bibitem{16}
P.~W. Battaglia, J.~B. Hamrick, and J.~B. Tenenbaum, ``Simulation as an engine
  of physical scene understanding,'' \emph{Proceedings of the National Academy
  of Sciences}, vol. 110, no.~45, pp. 18\,327--18\,332, 2013.

\bibitem{18}
K.~A. Smith and E.~Vul, ``Sources of uncertainty in intuitive physics,''
  \emph{Topics in Cognitive Science}, vol.~5, no.~1, pp. 185--199, 2013.

\bibitem{19}
P.~J. {Werbos}, ``Neural networks for control and system identification,'' in
  \emph{Proceedings of the 28th IEEE Conference on Decision and Control}, Dec
  1989, pp. 260--265 vol.1.

\bibitem{Atkeson97acomparison}
C.~G. Atkeson and J.~C. Santamaria, ``A comparison of direct and model-based
  reinforcement learning,'' in \emph{International Conference on Robotics and
  Automation}, 1997, pp. 3557--3564.

\bibitem{kurutach2018modelensemble}
T.~Kurutach, I.~Clavera, Y.~Duan, A.~Tamar, and P.~Abbeel, ``Model-ensemble
  trust-region policy optimization,'' in \emph{International Conference on
  Learning Representations}, 2018.

\bibitem{23}
M.~Pecka and T.~Svoboda, ``Safe exploration techniques for reinforcement
  learning -- an overview,'' in \emph{Modelling and Simulation for Autonomous
  Systems}, J.~Hodicky, Ed., 2014, pp. 357--375.

\bibitem{24}
J.~Garc{{\'i}}a and F.~Fern{{\'a}}ndez, ``A comprehensive survey on safe
  reinforcement learning,'' \emph{Journal of Machine Learning Research},
  vol.~16, pp. 1437--1480, 2015.

\bibitem{25}
S.~P. Coraluppi and S.~I. Marcus, ``Risk-sensitive and minimax control of
  discrete-time, finite-state markov decision processes,'' \emph{Automatica},
  vol.~35, no.~2, pp. 301 -- 309, 1999.

\bibitem{DBLP:journals/corr/abs-1109-2147}
P.~Geibel and F.~Wysotzki, ``Risk-sensitive reinforcement learning applied to
  control under constraints,'' \emph{J. Artif. Int. Res.}, vol.~24, pp.
  81--108, 2005.

\bibitem{26}
A.~Tamar, S.~Mannor, and H.~Xu, ``Scaling up robust mdps using function
  approximation,'' in \emph{Proceedings of the 31st International Conference on
  Machine Learning}, vol.~32, 2014, pp. 181--189.

\bibitem{8651519}
C.~D. {McKinnon} and A.~P. {Schoellig}, ``Learn fast, forget slow: Safe
  predictive learning control for systems with unknown and changing dynamics
  performing repetitive tasks,'' \emph{IEEE Robotics and Automation Letters},
  vol.~4, no.~2, pp. 2180--2187, 2019.

\bibitem{2}
V.~Mnih, K.~Kavukcuoglu, D.~Silver, A.~A. Rusu, J.~Veness, M.~G. Bellemare,
  A.~Graves, M.~Riedmiller, A.~K. Fidjeland, G.~Ostrovski, S.~Petersen,
  C.~Beattie, A.~Sadik, I.~Antonoglou, H.~King, D.~Kumaran, D.~Wierstra,
  S.~Legg, and D.~Hassabis, ``Human-level control through deep reinforcement
  learning,'' \emph{Nature}, vol. 518, no. 7540, pp. 529--533, Feb. 2015.

\bibitem{Garcia:2012:SES:2444851.2444864}
J.~Garc\'{\i}a and F.~Fern\'{a}ndez, ``Safe exploration of state and action
  spaces in reinforcement learning,'' \emph{J. Artif. Int. Res.}, vol.~45, pp.
  515--564, 2012.

\bibitem{safe}
D.~Sadigh and A.~Kapoor, ``Safe control under uncertainty with probabilistic
  signal temporal logic,'' in \emph{Robotics: Science and Systems XII}, 2016.

\bibitem{GARCIA1989335}
C.~E. García, D.~M. Prett, and M.~Morari, ``Model predictive control: Theory
  and practice—a survey,'' \emph{Automatica}, vol.~25, no.~3, pp. 335 -- 348,
  1989.

\bibitem{Borrelli:2017:PCL:3164811}
F.~Borrelli, A.~Bemporad, and M.~Morari, \emph{Predictive Control for Linear
  and Hybrid Systems}, 1st~ed.\hskip 1em plus 0.5em minus 0.4em\relax New York,
  NY, USA: Cambridge University Press, 2017.

\bibitem{1}
R.~S. Sutton and A.~G. Barto, \emph{Introduction to Reinforcement Learning},
  1st~ed.\hskip 1em plus 0.5em minus 0.4em\relax Cambridge, MA, USA: MIT Press,
  1998.

\bibitem{wunder2010classes}
M.~Wunder, M.~L. Littman, and M.~Babes, ``Classes of multiagent {Q}-learning
  dynamics with epsilon-greedy exploration,'' in \emph{Proceedings of the 27th
  International Conference on Machine Learning}, 2010, pp. 1167--1174.

\bibitem{mnih-atari-2013}
V.~Mnih, K.~Kavukcuoglu, D.~Silver, A.~Graves, I.~Antonoglou, D.~Wierstra, and
  M.~Riedmiller, ``Playing atari with deep reinforcement learning,'' in
  \emph{NIPS Deep Learning Workshop}, 2013.

\end{thebibliography}

\end{document}